# Analysis on Multi-robot Relative 6-DOF Pose Estimation Error Based on UWB Range

Xinran Li[a, b], Shuaikang Zheng[a], Pengcheng Zheng[a, b], Haifeng Zhang[a, b], Zhitian Li[a, *] and Xudong Zou[a, *]

*Abstract*—Relative pose estimation is the foundational requirement for multi-robot system, while it is a challenging research topic in infrastructure-free scenes. In this study, we analyze the relative 6-DOF pose estimation error of multi-robot system in GNSS-denied and anchor-free environment. An analytical lower bound of position and orientation estimation error is given under the assumption that distance between the nodes are far more than the size of robotic platform. Through simulation, impact of distance between nodes, altitudes and circumradius of tag simplex on pose estimation accuracy is discussed, which verifies the analysis results. Our analysis is expected to determine parameters (e.g. deployment of tags) of UWB based multi-robot systems.

## I. INTRODUCTION

The relative pose estimation is the foundational requirement for multiple-robot system. The existing multi-node team positioning technology often relies on the global navigation satellite system [1, 2]. In indoor or urban outdoor scenes, GNSS signals are blocked, resulting in weakened signals or undetectable signals, which cannot meet the requirements of positioning. In the scene without GNSS, Ultra Wide Band (UWB) technology has been paid increasing attention in relative localization research because of its low power consumption, high ranging accuracy and high security. To imply the multiple-robot system in open site, fixed anchors should be got rid of. Each robot is equipped with UWB tags. However, degeneracy (i.e. uncertainty of pose caused by the deployment of tags) is one of the main reason limit the scenes in which UWB based system can be applied. In [3], single-range based localization approaches are analyzed, concluding that in some cases, linear motion for example, relative motion between the robots cannot be observable.

To solve the degeneracy, multi-tag UWB system are proposed. In [4], each agent was equipped with two tags and a 9-axis Inertial Measurement Units, and sufficient condition for the observability of relative positions are given. In [5], a quadrotor flying with 4 tags was used to track a person. In [6], four anchor nodes were mounted on a flat mobile platform to positioning an Unmanned Aerial Vehicles with one tag.

Xinran Li contributes to this work
[a]The State Key Laboratory of Transducer Technology, Aerospace Information Research Institute, Chinese Academy of Sciences, Beijing 100190, China
[b]School of Electronic, Electrical and Communication Engineering, University of Chinese Academy of Sciences, Beijing 100049, China
*Corresponding author. E-mail addresses: ztli@mail.ie.ac.cn (Z. Li), zouxd@aircas.ac.cn (X. Zou).

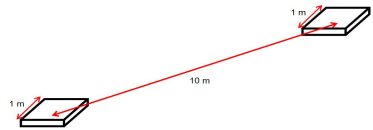

Fig. 1. Size of robots and distance between robots (example)

Although these methods significantly improve the degeneracy, some issues are still left behind. In [7], it is found that the height of the nodes and orientation cannot be estimated accurately in 3D case, which limiting the structure of tight-coupled systems and the freedom of motion. Experiments in [8] indicate that the uncertainty of position estimation is influenced by the deployment of tags. However, empirical proposals in reference [8] cannot provide enough guidance to the deployment of the system in lack of analysis. Moreover, experiments did not reveal the influence of deployment of tags on orientation estimation.

Hence, an analytical relationship between the uncertainty of the 6-DOF relative pose estimation and deployment of the multi-robot system should be derived, which can be implied to improve the accuracy of pose estimation in scenes with degeneracy and low accuracy. In particular, this paper describe the scenes where the distance between any two of the robots is far more than that between tags on a same robot, as Fig.1 shows.

The main contributions of this paper are summarized as:

1) An analytical lower bound of pose estimation error is given by analyzing the smallest eigenvalue of Jacobian matrix.

2) By analyzing the measurement model of Time of Arrival (TOA) UWB, the minimum configuration of the 3D relative pose estimation ranging system is given. By analyzing the Cramér-Rao Lower Bound, it is concluded that regular tetrahedron (triangle) is optimal when its circumradius or volume (size) is constrained.

3) Simulation experiments verify the analysis and demonstrate that when the other two factors are given, the pose estimation error can be linearly estimated by the distance, while its inverse can be linearly estimated by the altitude of tag simplex. The conclusion demonstrate linear approximate or interpolation is effective to determine the deployment of tags.

## II. PRELIMINARIES

### A. Notations

Without loss of generality, the relative positions between two nodes will be discussed. The conclusions can be extended o systems with any number of nodes. Select the local frame of a node as the global frame in this paper, abbreviated as FG. The node is called host node. The local frame of the other node, called tracked node, is denoted as FB. Then the transformation from coordinates under FB to coordinates under FG can be represented by the transformation matrix $T_{GB}$

$$\mathbf{T}_{GB} = \begin{bmatrix} \mathbf{C}_{GB} & \mathbf{t}_G^{BG} \\ \mathbf{0} & 1 \end{bmatrix} \in R^{4\times 4}, \mathbf{C}_{GB} \in R^{3\times 3}, \mathbf{t}_G^{BG} \in R^3 \quad (2\text{-}1)$$

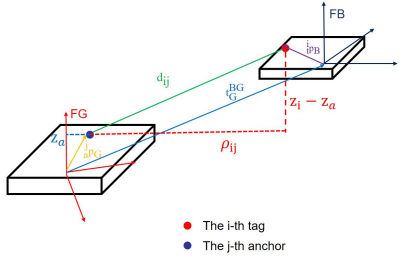

Fig. 2. System composition diagram

Where $\mathbf{C}_{GB}$ represents the rotation from FB to FG and $\mathbf{t}_G^{BG}$ is the position of FB under FG. $\mathbf{T}_{GB}$ can be parameterized to:

$$\boldsymbol{\theta} = [\mathbf{t}^T, \boldsymbol{\varphi}^T]^T \quad (2\text{-}2)$$

Where $\mathbf{t} := \mathbf{t}_G^{BG}$. In the 6 degree of freedom (DOF) case, $\boldsymbol{\varphi} \in R^3$ is a Lie algebra. In four degrees of freedom (3 translations, 1 heading Angle), $\varphi \in R$.

In order to distinguish UWB on both nodes, "anchor" will be used below to represent the UWB tag on the host node unless otherwise specified. Then the i-th ($i \in \{1, 2... m\}$) UWB tag and j-th ($j \in \{1, 2... n\}$) the distance of the anchor is expressed as

$$d_{ij} = \left\| \mathbf{t}_G^{BG} + \mathbf{C}_{GB}{}_t^i\mathbf{p}_B - {}_a^j\mathbf{p}_G \right\|_2 \quad (2\text{-}3)$$

Where ${}_t^i\mathbf{p}_B$ is the coordinate of tag i in FB and ${}_a^j\mathbf{p}_G$ is the coordinate of tag j under FG. The main symbols are shown in Fig. 2.

*B. Fisher Information Matrix*

Let $\mathbf{f}(\boldsymbol{\theta})$ and $\hat{\mathbf{d}}$ be noiseless and noisy distance vectors respectively:

$$\mathbf{f}(\boldsymbol{\theta}) = [d_{11}, d_{12} \cdots d_{mn}]^T \quad (2\text{-}4)$$

$$\hat{\mathbf{d}} = [\widehat{d_{11}}, \widehat{d_{12}} \cdots \widehat{d_{mn}}]^T \quad (2\text{-}5)$$

Suppose that $\hat{\mathbf{d}}$ follows a normal distribution of mean value $\mathbf{f}(\boldsymbol{\theta})$ and covariance matrix $\Sigma$, where $\Sigma = \sigma_d^2 \mathbf{I}_{mn\times mn}$. Let $\hat{\boldsymbol{\theta}}$ be an unbiased estimator of $\boldsymbol{\theta}$. Then the Cramér-Rao Lower Bound (CRLB) gives the lower bound of the covariance matrix represented by the Jacobian matrix $\mathbf{J}$:

$$E\left[(\boldsymbol{\theta} - \hat{\boldsymbol{\theta}})(\boldsymbol{\theta} - \hat{\boldsymbol{\theta}})^T\right] \geq (\mathbf{J}^T \boldsymbol{\Sigma}^{-1} \mathbf{J})^{-1} \quad (2\text{-}6)$$

$$\mathbf{J} = \frac{\partial \mathbf{f}(\boldsymbol{\theta})}{\partial \boldsymbol{\theta}} \quad (2\text{-}7)$$

CRLB is often used to compare the mean square error (MSE) of an estimate to the true value. In 6 DOF case, the i-th ($i = 1, 2,..., 6$) largest eigenvalues of the CRLB defined in (2-6) can be expressed as $\frac{\sigma_d^2}{s_{7-i}^{\mathbf{J}}{}^2}$, where $s_j^{\mathbf{J}}$ is the j-th largest right singular value of $\mathbf{J}$. Therefore, the lower bound of the estimation error can be given by estimating the singular value of $\mathbf{J}$.

*C. The Trilateration Method of Nonlinear Least Squares*

The position of the target point p is determined by reference points located at $\mathbf{p}_a^i$ ($i \in \{1, 2... m\}$) and their distance $d_i$. The nonlinear least squares method (NLLS) is obtained by minimizing the quasi-likelihood function

$$\underset{\mathbf{p}}{\operatorname{argmin}} \sum_i (d_i - \|\mathbf{p} - \mathbf{p}_a^i\|_2)^2 \quad (2\text{-}8)$$

The estimator $\hat{\mathbf{p}}$ obtained with the NLLS is optimal.

The NLLS in this paper will be used to solve the coordinate transformation from FB to FG, that is, to minimize the problem using the Levenberg-Marquardt (L-M) method:

$$\underset{\mathbf{C}_{GB},\ \mathbf{t}_G^{BG}}{\operatorname{argmin}} \sum_{i,j} (d_{ij} - \|\mathbf{t}_G^{BG} + \mathbf{C}_{GB}{}_t^i\mathbf{p}_B - {}_a^j\mathbf{p}_G\|_2)^2 \quad (2\text{-}9)$$

In Section IV, we use $\mathbf{t}_G^{BG}$ and $\boldsymbol{\varphi}$ to iterate, where $\boldsymbol{\varphi} \in R^3$ is the Lie algebra of $\mathbf{C}_{GB}$.

*D. Sigular Value of Matrix*

By the definition of singular value, it holds that for any $\mathbf{J} \in R^{m\times n}$ and $\mathbf{x} \in R^n$,

$$s_n^{\mathbf{J}} \leq \|\mathbf{J}\mathbf{x}\|_2 \quad (2\text{-}10)$$

Where $s_k^{\mathbf{J}}$ is the k-th largest right singular value of $\mathbf{J}$.

Another lemma is that the singular values is the continuous function of the elements of a finite matrix. We will give a brief proof.

The matrix $\mathbf{J}$ can be identified with the corresponding point in the m×n dimension Euclidean space. For any $\mathbf{B} \in R^{m\times n}$, where the Frobenius norm $\|\mathbf{B}\|_F = 1$, by definition

$$s_k^{\mathbf{J}+\mu\mathbf{B}} = \sqrt{\lambda_k(\mathbf{J}^T\mathbf{J} + \mu^2\mathbf{B}^T\mathbf{B} + \mu\mathbf{B}^T\mathbf{J} + \mu\mathbf{J}^T\mathbf{B})} \quad (2\text{-}11)$$

It holds that

$$\sqrt{\lambda_k(\mathbf{J}^T\mathbf{J}) - 2\mu s_1^{\mathbf{J}}} \leq s_k^{\mathbf{J}+\mu\mathbf{B}} \leq \sqrt{\lambda_k(\mathbf{J}^T\mathbf{J}) + \mu^2 + 2\mu s_1^{\mathbf{J}}} \quad (2\text{-}12)$$

If $s_1^{\mathbf{J}}$ is finite, $\lim_{\mu \to 0} s_k^{\mathbf{J}+\mu\mathbf{B}} = s_k^{\mathbf{J}}$.

III. THEORETICAL ANALYSIS

In this section, the measurement model of UWB and approximation of the right singular value of the Jacobian matrix defined in (2-9) is analyzed.

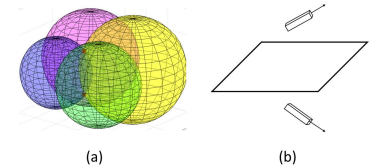

Fig. 3. Pose ambiguity

## A. Measurement Model of Anchors

Assume that there are n (n ≥ 3) spheres with coplanar centers in space. It is handy to prove there is an ambiguity of pose symmetric about the plane, as shown in Fig. 3 (b).

The pose ambiguity brought by coplanar anchors makes the precision of pose estimation dependent on the selection of initial values or the motion constraints. Therefore, at least 4 non-coplanar anchors are required in the UWB positioning system to avoid the above ambiguity.

## B. FIM Analysis on Anchor Placement

The pose ambiguity brought by coplanar anchors makes the precision of pose estimation dependent on the selection of initial values or the motion constraints. Therefore, at least 4 non-coplanar anchors are required in the UWB positioning system to avoid the above ambiguity.

Suppose there are n anchors and the parameter is the translation and rotation Lie algebra so(3), then the Jacobian can be expressed as the partitioned matrix:

$$\mathbf{J} = [\mathbf{J_t}, \mathbf{J_\varphi}] \quad (3\text{-}1)$$

Where the ((i-1)n + j)-th row of $\mathbf{J_t}$ is denoted as $\mathbf{u}_{ij}^T$, where

$$\mathbf{u}_{ij} = \frac{(\mathbf{t}_G^{BG} + \mathbf{C}_{GB}{}^i\mathbf{p}_B - {}^j_a\mathbf{p}_G)}{d_{ij}} \quad (3\text{-}2)$$

**Lemma 1**: The right singular value of $\mathbf{J_t}$ in (3-1) has the Euclidean transformation invariance of the global coordinate.

A brief explanation is given below: Let the global frame after the Euclidean transformation be FG ', and the coordinate of a point is **x** in FG and **x'** in FG', assuming **x'** = **Rx** + **t**. Suppose the first three columns of the Jacobian matrix in FG ' are $\mathbf{J_t}'$, its ((i-1)n + j)-th row is denoted by $\mathbf{u}_{ij}'^T$. Because $\mathbf{u}_{ij}'^T = \mathbf{u}_{ij}^T \mathbf{R}^T$, so $\mathbf{J_t}' = \mathbf{J_t}\mathbf{R}^T$. Since $\mathbf{R}^T$ is an orthogonal matrix, $\mathbf{J_t}$ and $\mathbf{J_t}'$ have the same right singular value.

First assume that the three anchors are in z=0 plane, the height of the 4-th anchor is set to $z_a > 0$, and the height of the i-th tag is set to $z_i$.

**Assertion**: By Section II-D, $\|\mathbf{J}_3\|_2$, the 2 norm of the third column of **J** is a good approximation of $s_6^J$ if for any i and j, $z_a \ll d_{ij}$, $z_i \ll d_{ij}$, and the distance between anchors and the distance between tags is much smaller than $d_{ij}$.

Suppose that the horizontal distance from the i-th tag to the j-th anchor $\rho_{ij}$, where $\rho_{ij} > \sqrt{3}\max(z_a, z_i)$ for any i and j. Take the partial derivative $\|\mathbf{J}_3\|_2^2$ against $z_t$, it can be seen that if the height of the i-th tag in FG satisfies

$$\frac{\rho_{i4}^2(z_i - z_a)}{(\rho_{i4}^2 + (z_i - z_a)^2)^2} + \sum_{j \neq 4} \frac{\rho_{ij}^2 z_i}{(\rho_{ij}^2 + z_i^2)^2} = 0 \quad (3\text{-}3)$$

$\|\mathbf{J}_3\|_2$ is an extrema. Because $\frac{d\|\mathbf{J}_3\|_2^2}{dz_i} < 0$ when $z_i < 0$ and $\frac{d\|\mathbf{J}_3\|_2^2}{dz_i} > 0$ when $z_i > z_a$, the solution of (3-3), represented as $z_i^0 = f_i(z_a)$, satisfies $0 < z_i^0 < z_a$. $\frac{d\|\mathbf{J}_3\|_2^2}{dz_i}$ monotonically increase when $0 < z_i^0 < z_a$, so there is a only solution of (3-3), which is the minimum point. Hence $\|\mathbf{J}_3\|_2$ is a function of $z_a$, represented as

$$J(f_i(z_a), z_a) \quad (3\text{-}4)$$

$\|\mathbf{J}_3\|_2$ increases when $z_i \in (-\infty, z_i^0]$ and decreases when $z_i \in [z_i^0, +\infty)$. Plug (3-3) in (3-2),

$$J(f_i(z_a), z_a)$$
$$= z_a \sqrt{\sum_i (\frac{1}{\sum_j \frac{\rho_{ij}^2}{d_{ij}^4}})^2 [\frac{1}{d_{i4}^2}(\sum_{j \neq 4} \frac{\rho_{ij}^2}{d_{ij}^4})^2 + (\frac{\rho_{i4}^2}{d_{i4}^4})^2 \sum_{j \neq 4} \frac{1}{d_{ij}^2}]} \quad (3\text{-}5)$$

By implicit function theorem and (3-3),

$$\frac{df_i(z_a)}{dz_a} = \sum_i \frac{\rho_{i4}^2 - 3[f_i(z_a) - z_a]^2}{\rho_{i4}^2 - 3[f_i(z_a) - z_a]^2 + \sum_{j \neq 4} \frac{\rho_{ij}^2 d_{i4}^6}{\rho_{i4}^2 d_{ij}^6}[\rho_{ij}^2 - 3f_i(z_a)^2]} \quad (3\text{-}6)$$

From (3-6), $0 < \frac{df_i(z_a)}{dz_a} < 1$ under the assumption that $\rho_{ij} > \sqrt{3}\max(z_a, z_i)$. Because

$$\frac{\partial}{\partial z_a} J^2(f_i(z_a), z_a)$$
$$= [\frac{\partial}{\partial z_i} J^2(f_i(z_a), z_a)\frac{df_i(z_a)}{dz_a} + \frac{\partial}{\partial z_a} J^2(f_i(z_a), z_a)]|_{z_i = f_i(z_a)} \quad (3\text{-}7)$$

It can be concluded that

$$\frac{\partial}{\partial z_a} J^2(f_i(z_a), z_a) = \sum_i [\frac{\rho_{i4}^2(z_a - f_i(z_a))}{(\rho_{i4}^2 + (f_i(z_a) - z_a)^2)^2}(1 - \frac{df_i(z_a)}{dz_a})$$
$$+ \sum_{j \neq 4} \frac{\rho_{ij}^2 f_i(z_a)}{(\rho_{ij}^2 + f_i(z_a)^2)^2} \frac{df_i(z_a)}{dz_a}] \quad (3\text{-}8)$$

Because $0 < f_i(z_a) < z_a$ and $0 < \frac{df_i(z_a)}{dz_a} < 1$ holds for any i, $J(f_i(z_a), z_a)$, the minimum of $\|\mathbf{J}_3\|_2$, is increased by $z_a$.

By Lemma 1, through translation and rotation, $z_a$ can be the length of any altitude of the anchor tetrahedron. Thus, it can be concluded that the shortest altitude of the anchor tetrahedron restricts the pose estimation accuracy.

*Remark 1.* Our analysis provides an estimation of the lower bound of the error. Denote d as the distance between the host node and tracked node and $z_a$ is the length of shortest altitude. If there are 3 tags and $z_a \ll d$, by (3-3) and (3-5), $z_i$ is approximately equal to $\frac{z_a}{4}$ and $J(f_i(z_a), z_a)$ is about $\frac{3z_a}{2d}$. Simulation results shows that when the following conditions are satisfied, the position estimation error can be amended to $C_1 \sigma_d \frac{d}{z_a} + D_1$, where $C_1$ and $D_1$ are constants when $z_a$ or d, the deployment of tags and $\sigma_d$ are determined.

(i) $\rho_{ij} > K_1 \max(z_a, z_i)$, where $K_1$ is a constant.

(ii) The altitudes of anchor tetrahedron and distance are the main factor limiting the estimation error (i.e. the minimum right singular value of $\mathbf{J_t}$ is a good approximation of $s_6^J$).

## C. Measurement Model of Tags

For each i ∈ {1, 2... m}, using $^j_a\mathbf{p}_G$ and $d_{ij}$ (j ∈ {1, 2... n}), the position of the i-th tag in FG $^i_t\mathbf{p}_G$ by (2-8), where $^i_t\mathbf{p}_G = \mathbf{t}^{BG}_G + \mathbf{C}_{GB}{}^i_t\mathbf{p}_B$. $\mathbf{C}_{GB}$ and $\mathbf{t}^{BG}_G$ can be solved by $^i_t\mathbf{p}_B$ and $^i_t\mathbf{p}_G$. If m = 2, there is at least one ambiguity of $\mathbf{C}_{GB}$ about rotating around the axis of the line between $^1_t\mathbf{p}_G$ and $^2_t\mathbf{p}_G$. If m = 3 and the 3 tags are not collinear, $\mathbf{t}^{BG}_G$ and $\mathbf{C}_{GB}$ have a unique solution

Therefore, in three-dimensional space, the UWB system needs at least 3 non-collinear tags to provide complete constraints for pose. However, if the motion is limited to 4DOF, then only two tags are needed, and line between them are not parallel to the axis of rotation.

## D. FIM Analysis on Tag Placement

In this section, the orientation estimation of nodes equipped with three non-collinear tags will be analyzed. In (3-1), the ((i-1)n + j)-th row of $\mathbf{J}_\varphi$ is

$$\mathbf{u}_{ij}{}^T \frac{\partial(\mathbf{C}_{GB}{}^i_t\mathbf{p}_B)}{\partial \varphi^T} \quad (3\text{-}11)$$

Different selection of parameters φ determines the different forms of $J_\varphi$. Suppose that any three-dimensional vector is $\mathbf{v} = [v_1, v_2, v_3]^T$, then $\mathbf{v}^\wedge = \begin{bmatrix} 0 & -v_3 & v_2 \\ v_3 & 0 & -v_1 \\ -v_2 & v_1 & 0 \end{bmatrix}$,

If $\varphi \in so(3)$, take an approximation of the derivative of $\mathbf{C}_{GB}{}^i_t\mathbf{p}_B$ against the Lie algebra

$$\frac{\partial(\mathbf{C}_{GB}{}^i_t\mathbf{p}_B)}{\partial \delta\varphi^T} \approx \mathbf{C}_{GB}({}^i_t\mathbf{p}_B)^\wedge \quad (3\text{-}12)$$

where $\delta\boldsymbol{\varphi}$ is the orientation error state [9], which is equivalent to $\frac{\partial(\mathbf{C}_{GB}{}^i_t\mathbf{p}_B)}{\partial \varphi^T}$ [10].

*Remark 2.* This approximation differs from other Lie algebra perturbing models by at most one right multiplication Jacobian matrix related to φ. Reference [11] gives detailed proof of this approximation.

By inverse lemma from matrix,

$$(\mathbf{J}^T\mathbf{J})^{-1} = \begin{bmatrix} * & * \\ * & \{\mathbf{J}_\varphi{}^T[\mathbf{I} - \mathbf{J}_t(\mathbf{J}_t{}^T\mathbf{J}_t)^{-1}\mathbf{J}_t{}^T]\mathbf{J}_\varphi\}^{-1} \end{bmatrix} \quad (3\text{-}13)$$

\* For unfocused elements.

Let the CRLB of the covariance matrix of $\widehat{\varphi}$, the estimator of φ, Define $\mathbf{H}_\varphi \triangleq \mathbf{J}_\varphi{}^T[\mathbf{I} - \mathbf{J}_t(\mathbf{J}_t{}^T\mathbf{J}_t)^{-1}\mathbf{J}_t{}^T]\mathbf{J}_\varphi$

$$E[(\varphi - \widehat{\varphi})(\varphi - \widehat{\varphi})^T] = (\mathbf{H}_\varphi)^{-1} \quad (3\text{-}14)$$

By eigenvalue decomposition,

$$\mathbf{I} - \mathbf{J}_t(\mathbf{J}_t{}^T\mathbf{J}_t)^{-1}\mathbf{J}_t{}^T = \mathbf{Q}\mathbf{D}\mathbf{Q}^T \quad (3\text{-}15)$$

Where the first three columns of $\mathbf{Q}$ belong to col($\mathbf{J}_t$) and $\mathbf{D}$ is the diagonal matrix, whose first three diagonal elements are 0 and the remaining diagonal elements are 1. Partition $\mathbf{J}_\varphi$ as

$$\mathbf{J}_\varphi \triangleq \mathbf{J}_\varphi{}^\perp + \mathbf{J}_\varphi{}^{//} \quad (3\text{-}16)$$

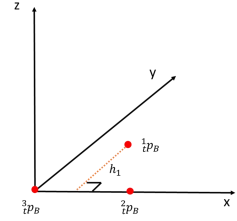

Fig. 4. Tags and altitude in FB

where $\mathbf{J}_\varphi{}^{//} \in \text{col}(\mathbf{J}_t)$ and $\mathbf{J}_\varphi{}^\perp$ is perpendicular to col($\mathbf{J}_t$).

The minimum eigenvalue of $\mathbf{H}_\varphi$ is $\lambda_3(\mathbf{H}_\varphi)$. Then

$$\lambda_3(\mathbf{H}_\varphi) \leq \left\|\mathbf{J}_\varphi{}^\perp{}_1\right\|_2^2 \left\|\mathbf{I} - \mathbf{J}_t(\mathbf{J}_t{}^T\mathbf{J}_t)^{-1}\mathbf{J}_t{}^T\right\|_2$$

$$= \left\|\mathbf{J}_\varphi{}^\perp{}_1\right\|_2^2 \leq \left\|\mathbf{J}_{\varphi_1}\right\|_2^2 \quad (3\text{-}17)$$

where $\mathbf{J}_\varphi{}^\perp{}_1$ and $\mathbf{J}_{\varphi_1}$ are respectively the first column of $\mathbf{J}_\varphi{}^\perp$ and $\mathbf{J}_\varphi$. The last inequality holds because of the orthogonality between $\mathbf{J}_\varphi{}^\perp$ and $\mathbf{J}_\varphi{}^{//}$.

**Lemma 2**: When the distance between the nodes is fixed, $\min_{\mathbf{T}_{GB}} \lambda_3(\mathbf{H}_\varphi)$ is has the Euclidean transformation invariance of the local coordinate FB.

For simplicity sake, we check only the 3 rows of the Jacobian corresponding to the different tags.

$$\mathbf{J}_\varphi = \begin{bmatrix} \mathbf{u}_1{}^T \mathbf{C}_{GB}({}^1_t\mathbf{p}_B)^\wedge \\ \mathbf{u}_2{}^T \mathbf{C}_{GB}({}^2_t\mathbf{p}_B)^\wedge \\ \mathbf{u}_3{}^T \mathbf{C}_{GB}({}^3_t\mathbf{p}_B)^\wedge \end{bmatrix} \quad (3\text{-}18)$$

If for i = 1, 2, 3, ${}^i_t\mathbf{p}_B' = \mathbf{R}{}^i_t\mathbf{p}_B + \mathbf{t}$, let $\mathbf{C}_{GB}' = \mathbf{C}_{GB}\mathbf{R}^T$. It can be concluded that $\mathbf{J}_t' = \mathbf{J}_t$. From (3-14),

$$\mathbf{J}_\varphi' = \mathbf{J}_\varphi\mathbf{R}^T + [\mathbf{u}_1 \; \mathbf{u}_2 \; \mathbf{u}_3]^T \mathbf{C}_{GB}\mathbf{R}^T(\mathbf{t})^\wedge \quad (3\text{-}19)$$

Thus $\min_{\mathbf{T}_{GB}} \lambda_3(\mathbf{H}_\varphi') = \min_{\mathbf{T}_{GB}\begin{bmatrix}\mathbf{R}^T & 0 \\ 0 & 1\end{bmatrix}} \lambda_3(\mathbf{R}\mathbf{H}_\varphi\mathbf{R}^T)$. Because $\mathbf{R} \in$ SO(3), $\min_{\mathbf{T}_{GB}} \lambda_3(\mathbf{H}_\varphi') = \min_{\mathbf{T}_{GB}} \lambda_3(\mathbf{H}_\varphi)$.

Define $h_i$ is the distance from the i-th tag to the line formed by the other 2 tags, as Fig. 4 shows.

After Euclidean transformation on FB, tags are in the plane where z = 0, the 2-nd and 3-rd tags lie on the x axis. $\mathbf{J}_{\varphi_1}$ can be represented as

$$\mathbf{J}_{\varphi_1} = [\mathbf{u}_{11} \; \mathbf{u}_{12} \; \mathbf{u}_{13} \; \mathbf{u}_{14} \; \mathbf{0}_{3\times 8}]^T \mathbf{C}_{GB} \begin{bmatrix} 0 \\ 0 \\ -h_1 \end{bmatrix} \quad (3\text{-}20)$$

$\mathbf{J}_3^{1:4}$ denotes the first four rows of the third column of $\mathbf{J}$.

$$\min_{T_{GB}} \lambda_3(\mathbf{H_\varphi}) \leq \min_{T_{GB}} \|\mathbf{J}_{\varphi_1}\|_2^2 \leq h_1^2 \|\mathbf{J}_3^{1:4}\|_2^2 \quad (3\text{-}21)$$

By Lemma 1 and Lemma 2, through translation and rotation in FG and FB, the shortest altitude of tag triangle and the anchor tetrahedron restricts the orientation estimation accuracy.

*Remark 3.* Denote d as the distance between the host node and tracked node and $h_1$ is the length of the shortest altitude of tag triangle. $z_a$ is defined in Remark 1. Suppose that for each j, $z_a \ll d_{1j}$, $\|\mathbf{J}_3^{1:4}\|_2^2$ is approximately equal to $\frac{3z_a^2}{4d^2}$. Simulation results shows that when the following conditions are satisfied, the maximum orientation error is approximately $C_2 \sigma_d \frac{d}{z_a h_1} + D_2$, where $C_2$ and $D_2$ are constants when $\sigma_d$ and two of $z_a$, $h_1$ and d are determined.

(i) $d_{ij} > K_2 z_a$, where $K_2$ is a constant.

(ii) The altitudes of anchor tetrahedron and distance are the main factor limiting the estimation error (i.e. $\|\mathbf{J}_{\varphi_1}\|_2$ is a good approximation of $\lambda_3(\mathbf{H_\varphi})$).

(iii) Because SO(3) is compact, the error of orientation cannot increase with the distance infinitely. $\sigma_d \frac{d}{z_a h_1}$ should not be too large to make the approximation and disturbance model in (3-12) false.

## IV. SIMULATION EXPERIMENTS

The simulation experiment was carried out in Matlab to analyze the factors affecting the accuracy of pose estimation obtained by ranging measurements when the nodes are in different relative poses. Add independent Gaussian noise with 5 cm standard deviation to the ranging measurements, i.e. $\sigma_d$ = 5 cm. Poses are estimated by NLLS. At each pose, 50 Monte-Carlo simulation experiments are carried out and the estimation error at this pose is calculated by the root-mean-square error (RMSE).

Positions of anchors and tags are shown in Fig. 5. According to the analysis in Section III, regular triangle and tetrahedron are optional when the volume (area) or circumradius of the anchor tetrahedron (tag triangle) is given, which is explained in Appendix.

The horizontal distance of tracked node is d and the altitude is z, as Fig.6 shows.

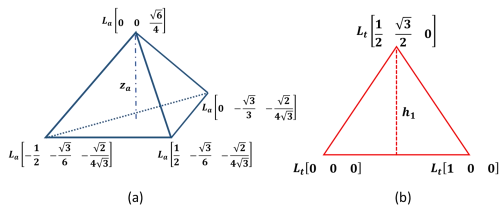

Fig. 5. (a) Anchors in FG (b) Tags in FB

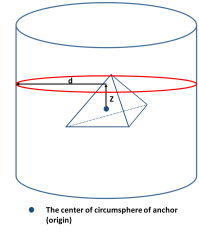

Fig. 6. Positions of tracked node

The error of translation is calculated by

$$E_t = \|\mathbf{t} - \hat{\mathbf{t}}\|_2 \quad (4\text{-}1)$$

where $\hat{\mathbf{t}}$ is the estimator of translation and t is the ground truth. The error of orientation is calculated by

$$E_\varphi = \|\ln(\mathbf{C_{GB}}^T \widehat{\mathbf{C_{GB}}})^\vee\|_2 \quad (4\text{-}2)$$

where $\widehat{\mathbf{C_{GB}}}$ is the estimator of rotation. $\ln(\mathbf{C_{GB}}^T \widehat{\mathbf{C_{GB}}})^\vee$ is the Lie algebra of $\mathbf{C_{GB}}^T \widehat{\mathbf{C_{GB}}}$.

For any fixed d, $L_a$ and $L_t$, maximum errors are taken in consideration, which can be represented by

$$\max_{T_{GB}} E_t \text{ (or } E_\varphi) \text{ s.t. horizontal distance is a constant} \quad (4\text{-}3)$$

### A. Error of Translation Estimatiom

Fig. 7 (b) and (c) shows that the altitude of tag triangle has little impact on error of translation estimation, while the translation error can be predicted linearly by distance between the nodes or the reciprocal of the altitude of anchor tetrahedron when the conditions in Remark 1 are satisfied.

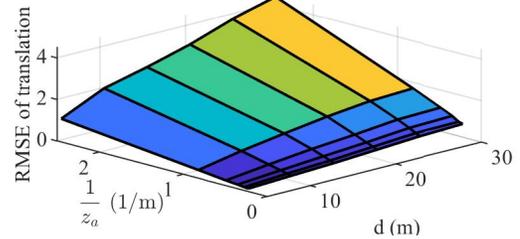

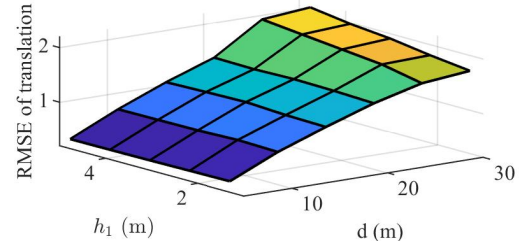

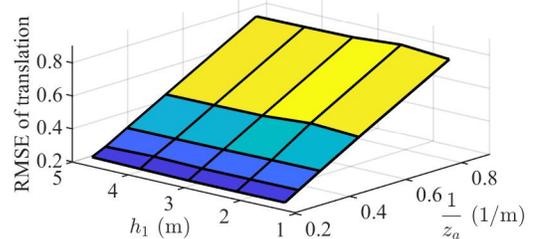

Fig. 7. Relationship between error of translation and other factors. In (a), (b) and (c), $L_t$, $L_a$ and d are respectively fixed at 1.5 m, 1.5 m and 10 m.

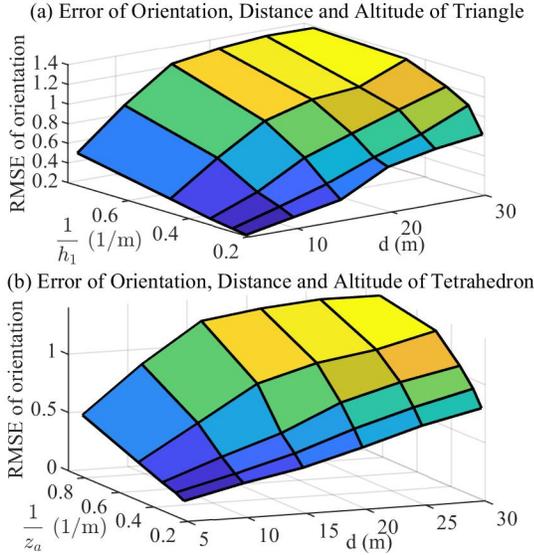

Fig. 8. Relationship between error of orientation and other factors. In (a) and (b), $L_t$ and $L_a$ are respectively fixed at 1.5m and 1.5m.

### B. Error of Orientation Estimatiom

Fig. 8 shows that the orientation error can be approximately predicted linearly by distance between the nodes or the reciprocal of the altitude of anchor tetrahedron and tag triangle when the conditions in Remark 3 are satisfied.

### C. Example

The requirement that the horizontal distance between the nodes is less than $d_{max}$ = 10 m. The RMSEs of translation and orientation are respectively less than $E_t$ = 0.5m and $E_\varphi$ = 0.3. It takes 2 steps to deploy the anchors and tags.

(i) Fix d = 10m and an adequately $L_t$ (Fig. 7 assures that when $L_t$ is adequately large, the translation error is independent of the tag placement.). RMSEs of translation with arbitrary two anchors satisfied conditions in Remark 1 can be used to determine the satisfactory $z_a$ and thus $L_a$. Through linear approximate (or interpolation) Fig. 9 (a), $L_a$ should be more than 2.5 m.

(ii) Fix d = 10 m and $L_a$ = 2.5 m. RMSEs of orientation with arbitrary two tags satisfied conditions in Remark 3 can be used to determine the satisfactory $h_1$ and $L_t$. Through linear approximate (or interpolation) in Fig. 9 (b), $L_t$ should be more than 3.2 m.

When $L_a$ = 2.5 m, $L_t$ = 3.2 m and d = 10 m, $E_t$ = 0.5067 m and $E_\varphi$ = 0.292.

## V. CONCLUSION

In this paper, an analytical lower bound of pose estimation error is given. Through simulation, impact of distance between nodes and altitudes of tags simplex on pose estimation is verified. Under the assumption that the distance between the nodes are far more than the size of robotic platform, A linear correction is found between the estimation error (or its inverse) and the factors, which can be used to determine the deployment of tags. Moreover, there are some areas to improve. One of the future direction is using timing analyze to figure out the pose estimation accuracy of UWB-IMU fusion system. Another direction is to find sufficient conditions for better relative pose performance, which requires to find out all degeneracy scenes and evaluate the error between CRLB and experimental results.

## APPENDIX

A tetrahedron is represented as $A_1A_2A_3A_4$, where $A_i$ is its node (i = 1, 2, 3, 4). The distance from $A_i$ to the opposite side is $h_i$. Let the volume of the tetrahedron be V and the circumradius be R, then

$$\sum_i h_i^2 \leq \frac{64}{9}R^2$$

$$\frac{\sqrt{3}}{8}\sqrt[3/4]{h_1h_2h_3h_4} \leq V$$

The equation holds if and only if $A_1A_2A_3A_4$ is a regular tetrahedron. [12]

A triangle is represented as $A_1A_2A_3$, where $A_i$ is its node and $h_i$ is its altitude. Let the area of the triangle be S and the circumradius be R, then

$$\sum_i h_i^2 \leq \frac{27}{4}R^2$$

$$\sqrt[2/3]{h_1h_2h_3} \leq \sqrt{3}S$$

The equation holds if and only if $A_1A_2A_3$ is a regular triangle.

## REFERENCES

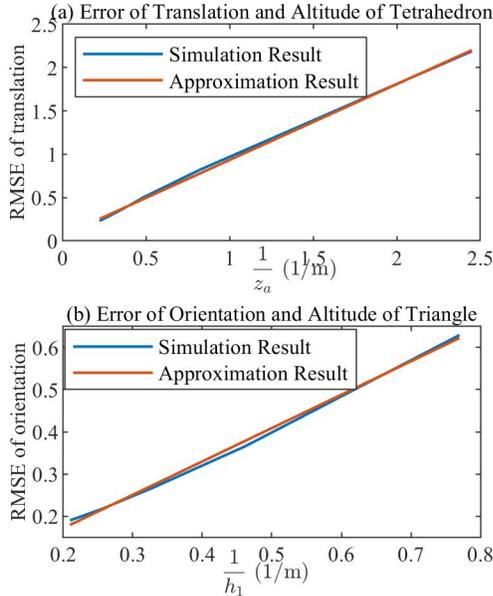

Fig. 9. Simulation for deployment of anchors and tags. In (a), d = 10m and $L_t$ = 1.5m. In (b), d = 10m and $L_a$ = 2.5m. The Pearson coefficient of correlation is respectively 0.9994 and 0.9986.